\documentclass{article}

\usepackage{arxiv}

\usepackage[utf8]{inputenc} 
\usepackage[T1]{fontenc}    
\usepackage{hyperref}       
\usepackage{url}            
\usepackage{booktabs}       
\usepackage{amsfonts}       
\usepackage{nicefrac}       
\usepackage{microtype}      
\usepackage{lipsum}
\usepackage{graphicx}
\graphicspath{ {./images/} }

\title{LLIA - Enabling Low-Latency Interactive Avatars: Real-Time Audio-Driven Portrait Video Generation with Diffusion Models}

\author{
  Haojie Yu\textsuperscript{*}\hspace{.5cm}Zhaonian Wang\textsuperscript{*}\hspace{.5cm}Yihan Pan\textsuperscript{*}\hspace{.5cm}Meng Cheng\hspace{.5cm}Hao Yang\hspace{.5cm} \\
  Chao Wang\hspace{.5cm}Tao Xie\hspace{.5cm}Xiaoming Xu$^\dagger$\hspace{.5cm}Xiaoming Wei\hspace{.5cm}Xunliang Cai \\[8pt]
  Meituan Inc. \\[8pt]
  \texttt{\{yuhaojie02, wangzhaonian, panyihan, chengmeng05, yanghao32, wangchao226,} \\
  \texttt{xietao13, xuxiaoming04, weixiaoming, caixunliang\}@meituan.com} \\[10pt]
  https://meigen-ai.github.io/llia/ \\
}

\begin{document}
\maketitle
\begin{abstract}
\footnotetext[1]{* Equal contributions.}
\footnotetext[2]{$\dagger$ Corresponding author.}
Diffusion-based models have gained wide adoption in the virtual human generation due to their outstanding expressiveness. However, their substantial computational requirements have constrained their deployment in real-time interactive avatar applications, where stringent speed, latency, and duration requirements are paramount. We present a novel audio-driven portrait video generation framework based on the diffusion model to address these challenges. Firstly, we propose robust variable-length video generation to reduce the minimum time required to generate the initial video clip or state transitions, which significantly enhances the user experience. Secondly, we propose a consistency model training strategy for Audio-Image-to-Video to ensure real-time performance, enabling a fast few-step generation. Model quantization and pipeline parallelism are further employed to accelerate the inference speed. To mitigate the stability loss incurred by the diffusion process and model quantization, we introduce a new inference strategy tailored for long-duration video generation. These methods ensure real-time performance and low latency while maintaining high-fidelity output. Thirdly, we incorporate class labels as a conditional input to seamlessly switch between speaking, listening, and idle states. Lastly, we design a novel mechanism for fine-grained facial expression control to exploit our model's inherent capacity. Extensive experiments demonstrate that our approach achieves low-latency, fluid, and authentic two-way communication. On an NVIDIA RTX 4090D, our model achieves a maximum of 78 FPS at a resolution of \( 384 \times 384 \) and 45 FPS at a resolution of \( 512 \times 512 \), with an initial video generation latency of 140 ms and 215 ms, respectively.
\end{abstract}

\section{Introduction}
\label{intro}
Portrait video generation\cite{prajwal2020lip, tian2024emo, xu2024vasa, zhang2023sadtalker} has become a key technology in a wide range of fields, including virtual assistants, digital entertainment, customer service solutions, and interactive education. By enabling the synthesis of realistic and expressive human portraits, this technology facilitates more immersive user experiences and supports creative content generation. The advent of diffusion models\cite{song2020denoising, ho2020denoising, blattmann2023stable} has further accelerated advances in portrait video generation\cite{tian2024emo}, providing a powerful framework for generating high-fidelity and temporally coherent videos.

More recently, GAN-based and diffusion model-based approaches have evolved in parallel, each paradigm offering unique advantages and facing distinct challenges. VASA-1\cite{xu2024vasa} develops a diffusion-based facial dynamics and head movement generation model that works in a face latent space, using a warping-based GAN as visual render\cite{drobyshev2022megaportraits} to achieve real-time performance. Further research\cite{zhu2024infp} extends for dyadic interaction, enable to dynamically alternate between speaking and listening states, guided by the input dyadic audio. Meanwhile, a number of methods\cite{chen2025echomimic,tian2024emo,xu2024hallo,cui2024hallo2,cui2024hallo3,wei2024aniportrait, lin2024cyberhost, jiang2024loopy} employing latent diffusion models have been proposed for portrait video animation. For instance, EMO\cite{tian2024emo} is the first audio-to-video framework to achieve highly expressive, seamless frame transitions, and identity preservation that requires only a single reference image and audio input. However, the advantages of each paradigm also highlight their respective limitations: the real-time efficiency of GANs often comes at the cost of limited expressiveness, whereas the impressive fidelity of diffusion models is typically hindered by high computational demands and slower inference speeds. To address the aforementioned challenges, we propose LLIA (low-latency interactive avatars), a portrait video generation framework based on the diffusion model, which serves as a render to achieve high-fidelity video generation.

Firstly, we introduce the concept of response latency in interactive digital avatars. During the initial response or state transitions, users must wait for the generation of a complete minimum-length video clip. Even if the digital avatar can be generated in real time, which means its inference FPS exceeds the video playback FPS, excessive response latency can still negatively impact the user experience. To reduce response latency, it is necessary to decrease the length of each generated video clip. However, empirically, generating clips that are too short can degrade the model's performance. Moreover, although the stable diffusion model inherently supports variable-length video generation, it tends to accumulate errors progressively when generating videos of lengths not encountered during training, ultimately leading to the collapse of the generated video content. Therefore, we propose a robust variable-length video generation training strategy that reduces the response latency while maintaining the quality of the generated videos.

Secondly, to address the computational challenges typically associated with diffusion models, we train consistency models\cite{song2023consistency} with DDPM pre-trained model to support few-step video generation. Building upon this, we further quantize the model into INT8 and process the time-intensive UNet and VAE modules concurrently to achieve an inference speed that enables real-time interaction with users. 

Thirdly, to enable effective interaction with users, the model should be able to behave naturally while waiting for instructions and also possess the ability to actively listen. Therefore, we categorize the states of the digital agent in conversational scenarios into three types: listening, idle, and speaking. Specifically, when in the idle state, the digital agent should remain motionless, awaiting further instructions. In the listening state, the agent should respond to the user's speech with encouraging gestures, such as nodding or smiling. To this end, we introduce class labels as conditional inputs to control the state of the digital avatar. By leveraging the acoustic features of the audio input, our digital agent demonstrates robust performance in interactive scenarios.

Lastly, we introduce explicit facial expression control capabilities into our framework to make digital humans more vivid and lifelike. Existing methods\cite{tian2024emo, xu2024hallo} deploy ReferenceNet\cite{hu2024animate} to provide appearance reference information, while they omit the disentanglement of appearance and expression. As a result, unintended facial expressions are transferred from the reference image to the generated video. Our method exploits this property by first applying a pre-trained portrait animation model\cite{guo2024liveportrait,xie2024x} to modify the expression in the reference image, thereby enabling the manipulation of facial expressions in the video. Taking advantage of the temporal continuity ensured by motion module\cite{guo2023animatediff}, our method enables explicit and seamless expression editing throughout the entire video sequence. This design significantly enhances the expressiveness and realism of the generated digital agents, supporting more engaging and interactive applications.

To support our training, we curate a high-quality dataset of over 100 hours by carefully selecting videos on the Web covering multiple topics and languages, followed by a rigorous screening process. In addition, we generated a batch of data in listening and idle states using portrait animation. The artificial data effectively maintains the generative ability and improves the performance in capturing listening and idle states.

Our main contributions can be summarized as follows:
\begin{itemize}
    \item We develop a robust variable-length video generation training paradigm specifically aimed at mitigating response latency in interactive digital avatars. Our model is capable of responding to users promptly while preserving high fidelity. When running on an RTX 4090D, our model produces digital avatars at \( 384 \times 384 \) resolution with a response latency of 140 ms, and at \( 512 \times 512 \) resolution with 215 ms.
    \item We propose an efficient consistency model training strategy specifically for the Audio-Image-to-Video task. During inference and deployment, we further quantized the model to INT8 and parallelized the UNet and VAE modules. Collectively, these methods enable our model to achieve real-time performance. On an RTX 4090D machine, the model generates digital avatars at \( 384 \times 384 \) resolution with 78 FPS, and at \( 512 \times 512 \) resolution with 45 FPS. 
    \item A method for long-term stable inference of specific portraits with very few steps has been designed in this work. Specifically, the diffusion process of the last four frames during inference for an idle state is replaced through a specific strategy, guiding the diffusion process to one of several predefined states to achieve stable generation.
    \item We design three conversational states for the model, which can be controlled via class labels, and fine-grained facial expression control. The combination of these two features enhances the user interaction experience.
    \item We introduce a high-quality dataset of over 100 hours, which comprises open-source data, web-collected data spanning various topics, and synthetic data specifically designed to improve the model’s performance in listening and idle states.
\end{itemize}

\section{Related work}
\paragraph{Non-diffusion-based Portrait Animation}
Early works in the field of digital humans focus primarily on video-based talking head generation, which allows direct editing of an input video segment. For example, Wav2Lip\cite{prajwal2020lip} proposes a novel network to sync the lip of arbitrary videos of talking faces with arbitrary speech in the wild. Subsequently, portrait animation techniques evolved from FOMM\cite{siarohin2019first} to implicit-keypoints-based methods\cite{guo2024liveportrait, drobyshev2022megaportraits}. These approaches employ implicit keypoints as intermediate motion representations and warp the source portrait using the driving image through optical flow. VASA-1\cite{xu2024vasa} adopts this technique as its rendering module and develops a Diffusion Transformer model that operates in the latent space of motion representations, enabling animation from a single image. INFP\cite{zhu2024infp} further advances the model for interactive scenarios using dyadic audio guidance. However, these methods rely on a warping-based GAN for visual rendering, which constrains the expressiveness of the virtual avatars.

\paragraph{Diffusion-based Portrait Video Generation}
With the advent of diffusion models, due to their superior diversity and enhanced controllability compared to traditional image generation techniques such as GANs or VAEs, numerous research efforts\cite{ho2020denoising, song2020denoising} have adopted diffusion models for image generation tasks. Pioneering works like AnimateDiff\cite{guo2023animatediff} and SVD\cite{blattmann2023stable} have successfully applied diffusion models to video generation. Building upon these advancements, significant progress\cite{tian2024emo, chen2025echomimic, jiang2024loopy, zhu2024infp, xu2024vasa, wei2024aniportrait, wang2025omnitalker, lin2024cyberhost, qi2025chatanyone} has been achieved in portrait video generation based on diffusion models.

Specifically, Animate Anyone\cite{hu2024animate} proposed a stable portrait video generation framework comprising ReferenceNet-DenoisingNet-Guider architectures. This paradigm enables the generation of customized video sequences depicting specific actions for target individuals, utilizing both reference portraits and action sequence images as input. The framework employs SD1.5 as pre-trained weights, integrates the motion module proposed in AnimateDiff\cite{guo2023animatediff} for temporal coherence, and implements Guider to process action maps, which are subsequently fused with noise maps and intermediate features. The ReferenceNet processes reference portraits and interacts with DenoisingNet's intermediate layers to provide stable facial feature guidance, while CLIP\cite{radford2021learning} contributes additional facial abstract information for enhanced generation quality.

Based on Animate Anyone and related work, the Alibaba team subsequently proposed EMO (Emote Portrait Alive)\cite{tian2024emo} for talking head generation controlled by portrait images, audio signals, and additional control information. While maintaining a similar architectural framework to Animate Anyone, EMO modifies the Guider input to utilize head motion range masks instead of action maps. To improve temporal continuity, ReferenceNet dynamically incorporates previous video frames as reference. The architecture innovatively introduces audio-attention layers to integrate speech information and implements a speed control bucket mechanism to regulate head motion velocity, thereby achieving highly expressive talking-head generation.

Subsequent to EMO's development, various derivative studies have emerged, including the Hallo series models\cite{xu2024hallo, cui2024hallo2, cui2024hallo3}. These extensions demonstrate the expanding research frontier in diffusion-based portrait video generation, with continuous technical innovations in temporal modeling, multi-modal integration, and motion controllability.

\section{Method}
\begin{figure} 
    \centering
    \includegraphics[scale=0.56]{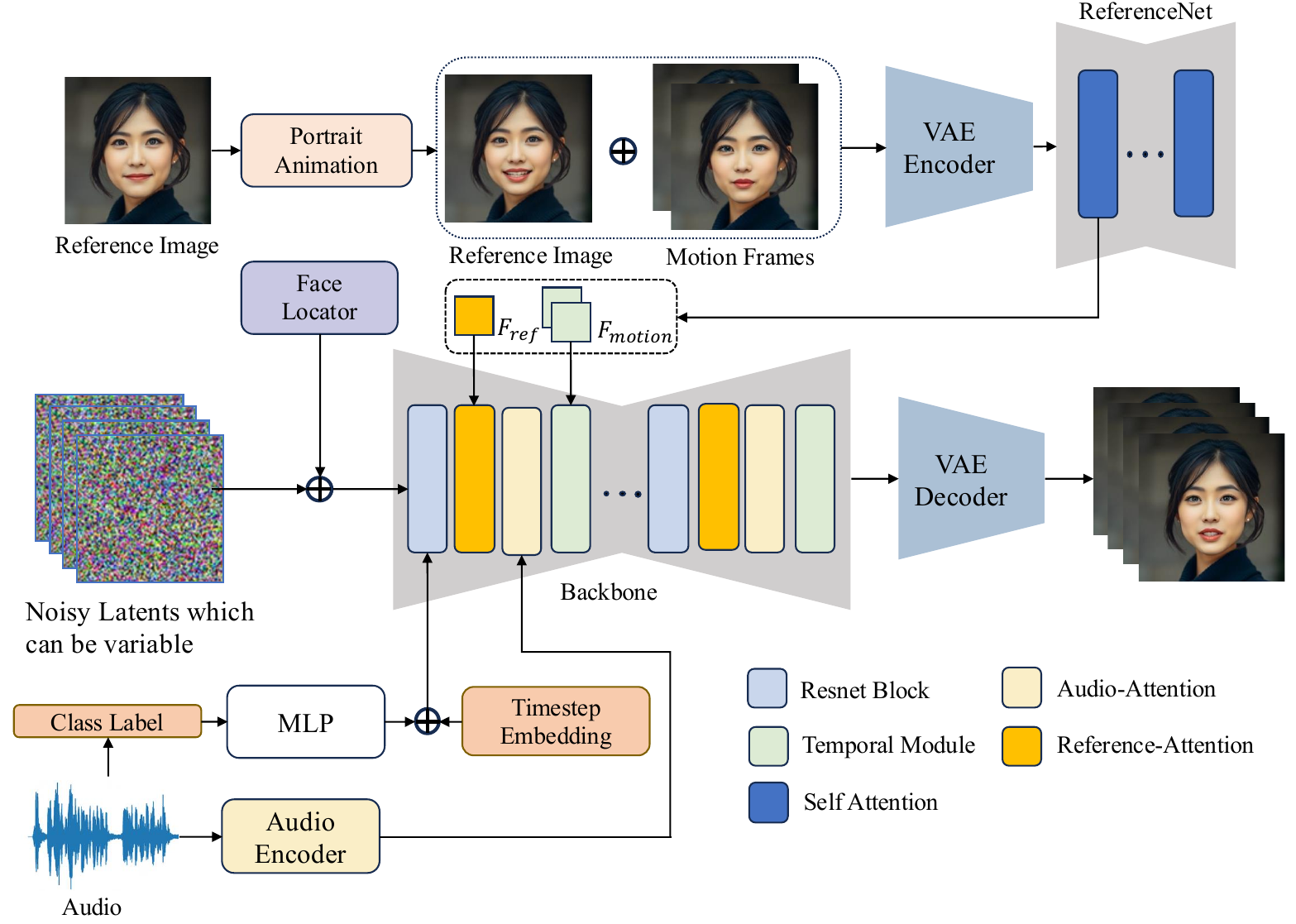}
    \caption{Overview of the proposed method. Compared with previous work, our pipeline introduces several novel modules: 1) before feeding the reference portrait into the ReferenceNet, we first apply portrait animation to adjust its facial expression to match our provided template; 2) class labels used to determine the avatar’s state are obtained from the input audio. These labels can be inferred directly from acoustic features, or alternatively, guided by an LLM model to indicate the appropriate state; 3) the length of the noisy latent is kept fixed during the early stage of training. In the later stage, it becomes dynamic, enabling the model to gain the capability of variable-length video generation.}
    \label{Overview}
\end{figure}

Our proposed method enables the generation of digital agent with natural head movements and facial expressions, given a single portrait image and an audio clip. By harnessing the capabilities of large language models\cite{hurst2024gpt}, our system supports real-time interactive conversations with users, seamlessly switching between listening, idle and speaking states. In addition, our approach allows for the dynamic modification of the digital agent's facial expressions, enhancing the overall user experience. The overview of the proposed method is shown in Figure \ref{Overview}.

\subsection{Network Pipelines}
Our model is built on the diffusion model framework, employing the same backbone architecture as Stable Diffusion 1.5 (SD1.5)\cite{blattmann2023stable}. We discard the use of prompt embedding and focus on alternative conditioning mechanisms. Similarly to previous work\cite{tian2024emo, hu2024animate}, our approach integrates four additional modules: 1) \textit{Appearance ReferenceNet}; 2) \textit{Audio Layer}; 3) \textit{Motion Module}; 4) \textit{Face Locator}. It is worth noting that given the requirement for streaming video generation, we extract audio segments with a predetermined length and process them through a pre-trained Wav2Vec\cite{baevski2020wav2vec, schneider2019wav2vec, TencentGameMate} to obtain audio features. To achieve minimal latency, for each frame's motion, we only input the audio features corresponding to the preceding frames. These features are then incorporated into the model via a cross-attention mechanism after reference-attention layers, enabling effective interaction between the audio and visual modalities. 

\subsection{Variable-length Video Generation}
As defined in Section \ref{intro}, it is necessary to minimize response latency in order to deliver more immediate feedback to users. Our empirical results indicate that the generation of longer video segments in a single batch enhances the overall performance and coherence of the model. However, producing extended video clips in one sampling also introduces higher delay and increases GPU memory requirements. To balance these trade-offs, we propose a training strategy for robust variable-length video generation. Previous works\cite{xu2024hallo, tian2024emo} sample $n+f$ contiguous frames from the video clip, with the started $n$ frames are motion frames. Both $n$ and $f$ are kept fixed during video training. However, we introduce a dynamic training strategy. Specifically, at the beginning of training, $f$ is kept constant to facilitate effective training of the motion module and audio layer, enabling the model to achieve satisfactory video generation performance. Once the model is sufficiently trained, we make $f$ dynamic during training, ensuring that $f \geq n$.  Due to the dynamic training, we can adopt different video generation lengths during inference without accumulating errors in subsequent frames. A shorter generation length is used first to significantly reduce the response latency. For the following segments, a longer generation length is maintained to ensure high video quality and coherence. Moreover, we can select an appropriate inference length based on the available GPU memory of the deployment device, allowing our model to be adaptable to a wider range of application scenarios.

\subsection{Class Labels}
To create a conversational agent capable of engaging in natural interactions, the model must not only focus on generating speech, but also effectively capture listening and idle states. More specifically, in the idle state, the digital agent should maintain a stationary posture, passively waiting for further instructions. Conversely, in the listening state, the agent ought to engage with the user's speech through encouraging nonverbal cues, such as nodding or smiling.

To this end, we introduce class labels to explicitly control the agent's state. We define three distinct states: speaking, listening, and idle. As these states are mutually independent, we simply use an embedding layer to encode class labels. The encoded class labels are subsequently added to the timestep embeddings. With a well-annotated dataset\ref{subsec:dataset}, our model learns to differentiate among these states and can switch roles by providing the corresponding class information during inference. The specific state signal can be determined based on the acoustic features of the user's audio input. Consequently, this approach enables the model to flexibly interpret the user's intent and generate appropriate state transitions.

\subsection{Expression Control}
Facial expression control is a crucial step in the development of digital agents, as it allows animations to be more expressive and can be used in different scenarios. VASA-1\cite{xu2024vasa} uses emotion coefficients extracted by \cite{savchenko2022hsemotion} to modulate the depicted emotion. Hallo2\cite{cui2024hallo2} incorporates semantic textual labels for portrait expressions as conditional inputs.

In our method, we propose an approach that allows explicit control over the facial expressions of digital agents. As mentioned above, our model leverages the ReferenceNet to maintain identity and background consistency. In the process, all facial information of the reference image, including its facial shape and expression, is incorporated into the model. In the self-supervised training paradigm, the model inevitably tends to copy the expression information from the reference image as a short-cut. Based on this property, before feeding the reference image into ReferenceNet, we introduce an additional portrait animation model\cite{guo2024liveportrait, xie2024x} to explicitly modify the expression of the reference image. We find that this approach effectively achieves expression manipulation. Due to the presence of the motion module, the entire modification process is natural and seamless, and does not introduce additional computational overhead during inference.

\subsection{Consistency Models}
\paragraph{Training Challenges and Solutions.}
Reducing the number of steps in diffusion models constitutes a highly effective acceleration strategy. Although existing research focuses primarily on T2I or T2V tasks, our framework incorporates multiple control conditions, making the reduction of inference steps particularly challenging. Initial experiments revealed that exclusive reliance on LCM loss resulted in ambiguity in facial structure and mild deformation artifacts. Inspired by previous work\cite{lin2024animatediff, lin2024sdxl, goodfellow2020generative}, we introduced discriminators to enhance visual quality. However, direct adoption of the 3D discriminator architecture from AnimatedDiff-Lightning proved suboptimal for our Audio-Image-to-Video (AI2V) task, which imposes stricter constraints than Text-to-Video (T2V) scenarios. The original implementation's excessive focus on facial details and inter-frame relationships led to degradation in lip synchronization accuracy and identity preservation. To address this, we augment the discriminator with audio features and reference network information, allowing updates to the audio cross-attention weights and head-related structural parameters during adversarial training.

\paragraph{Implementation Details.}
In pursuit of real-time generation, we adopted methodologies from prior works by incorporating Latent Consistency Models (LCM) and discriminators during the training phase to allow for fewer-step diffusion sampling\cite{wang2024animatelcm}. LCM operates by constructing latent space self-consistency constraints during training\cite{luo2023latent}, facilitating single/multi-step noise-to-image generation. For initialization, our model utilized weights pretrained through the denoising diffusion probabilistic models (DDPM) process. Throughout training, auxiliary structures, including the ReferenceNet, Face Locator, and audio feature extractors, solely provide reference features without participating in the construction of diffusion self-consistency constraints. Consequently, these components remain frozen and only the UNet and motion module parameters are updated.

\begin{equation}
D_{h}^{x_{*}} = \left\{ D_h\left( 
x_{*}, \, t, \, \mathbf{h}_{enc}, \, \mathbf{f}_{face}, \, \mathbf{a}_{audio}, \, \mathbf{h}_{motion} 
\right) \right\}_{h=1}^{H}
\end{equation}

\begin{equation}
\mathcal{L}_D = \frac{1}{H} \sum_{h=1}^{H} \mathbb{E}{(x_{fake}, x_{real})} \left[ w \cdot ReLU\left(D_h^{x_{fake}} + 1\right) + w \cdot ReLU\left(1 - D_h^{x_{real}}\right) \right]
\end{equation}

\begin{equation}
\mathcal{L}_G = \frac{1}{H} \sum_{h=1}^{H} \mathbb{E}_{x_{fake} \sim G} \left[ w \cdot ReLU\left(1 - D_{h}^{x_{fake}} \right) \right]
\end{equation}

\begin{equation}
\mathcal{L}_{lcm} = \mathop{\mathbb{E}}_{i,j} \left[ \left((y_{pred,i,j} - y_{target,i,j})^2 + c^2\right)^{\frac{1}{2}} - c \right ]
\end{equation}

\begin{equation}
\mathcal{L} = \lambda_0\mathcal{L}_{lcm} + \lambda_1\mathcal{L}_G
\end{equation}

The DDIM solver was used to establish adjacent point pairs in the Probability Flow ODE (PF-ODE) process\cite{song2020denoising}. Consistent with our training paradigm, we implemented v-prediction parameterization with zero SNR weighting, adopting a learning rate of 3e-6 and a 1000-step warm-up.

\paragraph{Temporal Consistency Optimization.}
Despite these improvements, the output videos exhibited temporal flickering artifacts. Using the high similarity between adjacent frames, we implemented a frame weighting strategy where predicted frames were temporally smoothed before calculating the reconstruction loss against target frames. Empirical validation confirmed that this approach effectively mitigated the flickering phenomenon.

\paragraph{Architectural Insights.}
In particular, we observed that training the motion module solely achieved performance comparable to joint optimization of the UNet and motion module. However, complete UNet fine-tuning remains essential for character-specific adaptation. Our optimal training protocol involves:

\begin{enumerate}
    \item Multi-scale single-frame pre-training.
    \item Subsequent video data training using these pre-trained weights (including discriminator parameters) to optimize motion dynamics.
\end{enumerate}

\paragraph{Training Paradigm Comparison.}
Contrary to knowledge distillation approaches, we empirically verified that direct training on target data yields more stable convergence and superior results. This paradigm demonstrates particular effectiveness in maintaining identity consistency while achieving real-time generation capabilities (less than four sampling steps).

\subsection{Quantization and Deployment}
To effectively deploy applications in the industry, we have conducted engineering optimizations for model deployment. These mainly include inference engine optimization, model quantization, and pipeline parallelism (latency hiding). Detailed acceleration results are presented in Table \ref{tab:Inference Latency}.

\paragraph{Inference Engine Optimization.}
In industrial deployment scenarios, NVIDIA GPUs are the most widely used. Therefore, we utilize NVIDIA inference engine TensorRT to accelerate model inference. TensorRT achieves efficient inference through techniques such as layer fusion, automatic kernel selection, and memory optimization. On traditional CNN models, TensorRT half-precision (FP16) generally offers 50\%\textasciitilde100\% speedup compared to PyTorch half-precision (FP16), with little change in quality. Since the VAE and UNet in our model are CNN-based models, we employ the TensorRT inference engine for acceleration.

\paragraph{Model Quantization.}
In our pipeline, the UNet model accounts for the majority of the inference time, more than 60\%. To reduce this inference time, we employed model quantization techniques on the UNet. We selected INT8 precision for quantization, as it imposes lower requirements on GPU hardware compared to FP8, which is only supported on SM89 and above. Furthermore, INT8 preserves accuracy better than other lower bits formats such as INT4. For 3D UNet models, direct quantization can lead to unsupported operators and may not yield noticeable acceleration. To address this, we utilized efficient operators from 2D implementations to equivalently support 3D UNet, ensuring satisfactory acceleration. Despite this, certain accuracy concerns remained, prompting us to identify and fine-tune sensitive layers. High precision inference was applied to these layers to maintain consistency between optimized and pre-optimized inference results. After a series of optimizations, INT8 quantization achieved significant speed enhancements while generally maintaining the original model performance.

\begin{figure} 
    \centering
    \includegraphics[scale=0.22]{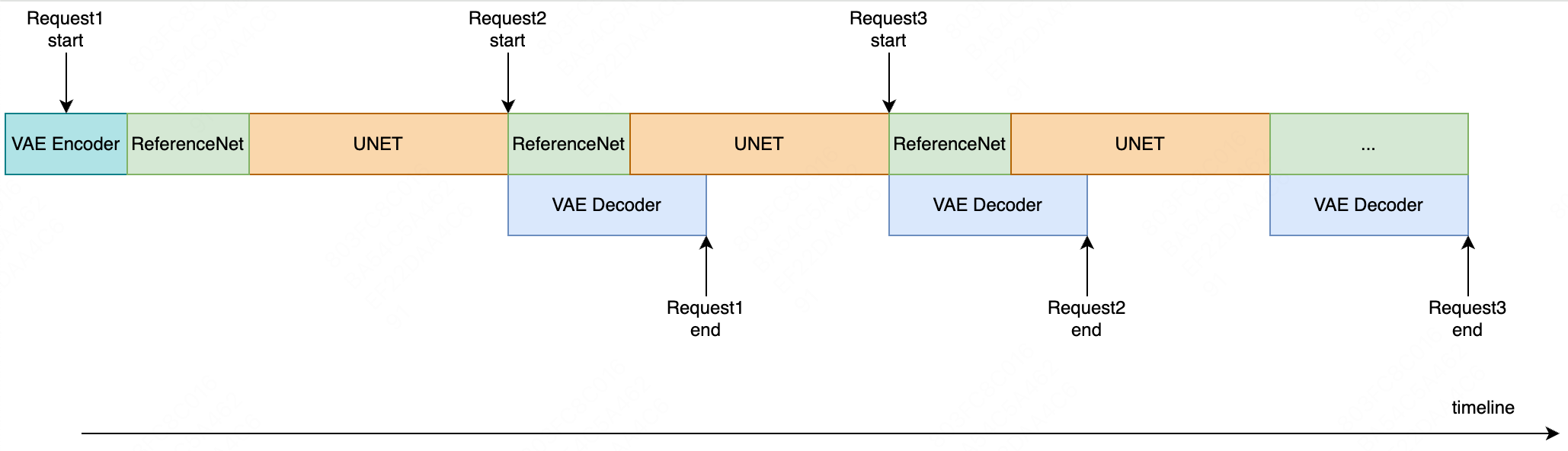}
    \caption{Illustration of Inference Pipeline Parallelism.}
    \label{Inference Pipeline}
\end{figure}

\paragraph{Pipeline Parallelism.}
The pipeline of this model consists of multiple modules. By leveraging pipeline parallelism, we can process the time-intensive UNet and VAE modules concurrently, effectively masking the execution time of certain modules as illustrated in Figure \ref{Inference Pipeline}. This strategy improves the request handling capacity and achieves higher throughput.

\begin{table}
 \caption{Comparison of Inference Latency Before and After Optimization. The experiments were conducted on an RTX 4090 machine with 4-step sampling, a resolution of $384 \times 384$, and 12 sample frames, under classifier-free guidance.}
  \centering
  \begin{tabular}{llll}
    \toprule
    Model Stage & Before Optimization & After Optimzation & Speedup Ratio \\
    \midrule
    VAE Encoder & 56.8 ms  & 31.7 ms &  1.79x    \\
    Denoise UNET & 880 ms & 418 ms & 2.10x  \\
    VAE Decoder & 202 ms & 104 ms & 1.94x \\
    \bottomrule
  \end{tabular}
  \label{tab:Inference Latency}
\end{table}

\section{Experiments}
\subsection{Dataset}
\label{subsec:dataset}
This study established a high-quality dataset for model training through systematic data curation. The dataset integrates open source resources (MEAD, VoxCeleb2 and CelebV-HQ)\cite{kaisiyuan2020mead, chung2018voxceleb2, zhu2022celebv} with curated videos on the Web, collectively exceeding 10,000 hours of raw video content. A rigorous multistage filtering pipeline was implemented to ensure data integrity.

The initial automated processing involved scene segmentation, facial detection, optical flow analysis, lip-sync verification, and head pose estimation for preliminary filtering\cite{li2023yolov6, lugaresi2019mediapipe}. Subsequent manual screening evaluated facial resolution quality, background stability, and speech activity patterns. Through this dual-phase curation process, the final training sets were refined to 171 hours for Stage I and 158 hours for Stage II model development. The methodology demonstrates effective quality control through technical verification and integration of human expertise.

In addition, considering the difficulty in collecting data for listening and idle states (the proportion remaining after processing online videos is less than 1\%), we generated a batch of such data using portrait animation\cite{guo2024liveportrait}. Specifically, we selected over one thousand representative identities from the previously collected videos, and combined them with carefully selected action segments, resulting in more than four hours of generated data. This data effectively preserves the generative capability of the model and enhances its performance in representing listening and idle states.

\subsection{Implement details}
\paragraph{Training Configuration.}
This study used 48 A100 GPUs for training. In Stage I, the model architecture described above was implemented with the following specifications: The model weights were initialized using Stable Diffusion v1.5 (SD1.5). During training, the input images were randomly scaled to resolutions of 256, 384, or 512 within each minibatch. The training configuration used a batch size per GPU of 16, AdamW optimizer with learning rate 1.0e-5, and a 200-step warm-up period. The beta schedule followed a scaled\_linear strategy with v\_prediction as the prediction target type, while enabling rescale\_betas\_zero\_snr.
The model was first trained for 100,000 steps using MSE loss, followed by an additional facial region loss weighting scheme implemented via MediaPipe for feature extraction. The total duration of the training was extended to 200,000 steps.

In Stage II, the motion module was initialized with AnimateDiff v2 weights. We used 4 motion frames and a dynamically varying generated video length ranging from 4 to 12 frames. To enhance Chinese lip synchronization, the audio encoder was replaced with Chinese-Wav2Vec2-Base\cite{baevski2020wav2vec, schneider2019wav2vec}. Training exclusively updated parameters in the motion module, audio projector, and audio cross-attention layers within UNet. This stage maintained the MSE loss and fixed target resolution training, with other hyperparameters identical to Stage I.

\paragraph{Accelerated Training Phases (denoted as extra Stage I and extra Stage II for distinction).}
During extra Stage I, only UNet weights and the discriminator head parameters were updated. The training framework incorporated a PF-ODE solver based on DDIM to construct adjacent point pairs. Both UNet and the discriminator used AdamW optimizer with learning rate 3e-6, where the discriminator updates began after 10,000 steps. The configuration included 1,000 warm-up steps, per-GPU batch size 8, and combined LCM-adversarial loss objectives. Other settings mirrored Stage I, with training lasting 80,000 steps.

Extra Stage II exclusively updated motion module parameters, audio cross-attention layers, and discriminator's audio/head components. A supplementary loss between frames (weight = 0.15) was introduced between the predicted frames and adjacent frames in the LCM loss computation. All remaining configurations matched Stage II specifications, and the training was extended to 50,000 steps.

This implementation demonstrates a hierarchical training strategy with phased parameter updates and specialized loss formulations to optimize both visual quality and audio-visual synchronization.

\begin{table}
 \caption{Inference Speed.}
  \centering
  \begin{tabular}{lllll}
    \toprule
    Sampling steps & Resolution & Sample frames & Pipe (ms) & DenoisingNet (ms) \\
    \midrule
    2 & $384 \times 384$ & 4 & 100 & 48 \\
    4 & $384 \times 384$ & 4 & 151 & 95 \\
    2 & $512 \times 512$ & 4 & 147 & 81 \\
    4 & $512 \times 512$ & 4 & 233 & 161 \\
    2 & $384 \times 384$ & 12 & 154 & 96 \\
    4 & $384 \times 384$ & 12 & 258 & 192  \\
    2 & $512 \times 512$ & 12 & 265 & 189  \\
    4 & $512 \times 512$ & 12 & 469 & 376  \\
    \bottomrule
  \end{tabular}
  \label{tab:Inference Speed}
\end{table}

\begin{figure} 
    \centering
    \includegraphics[scale=0.6]{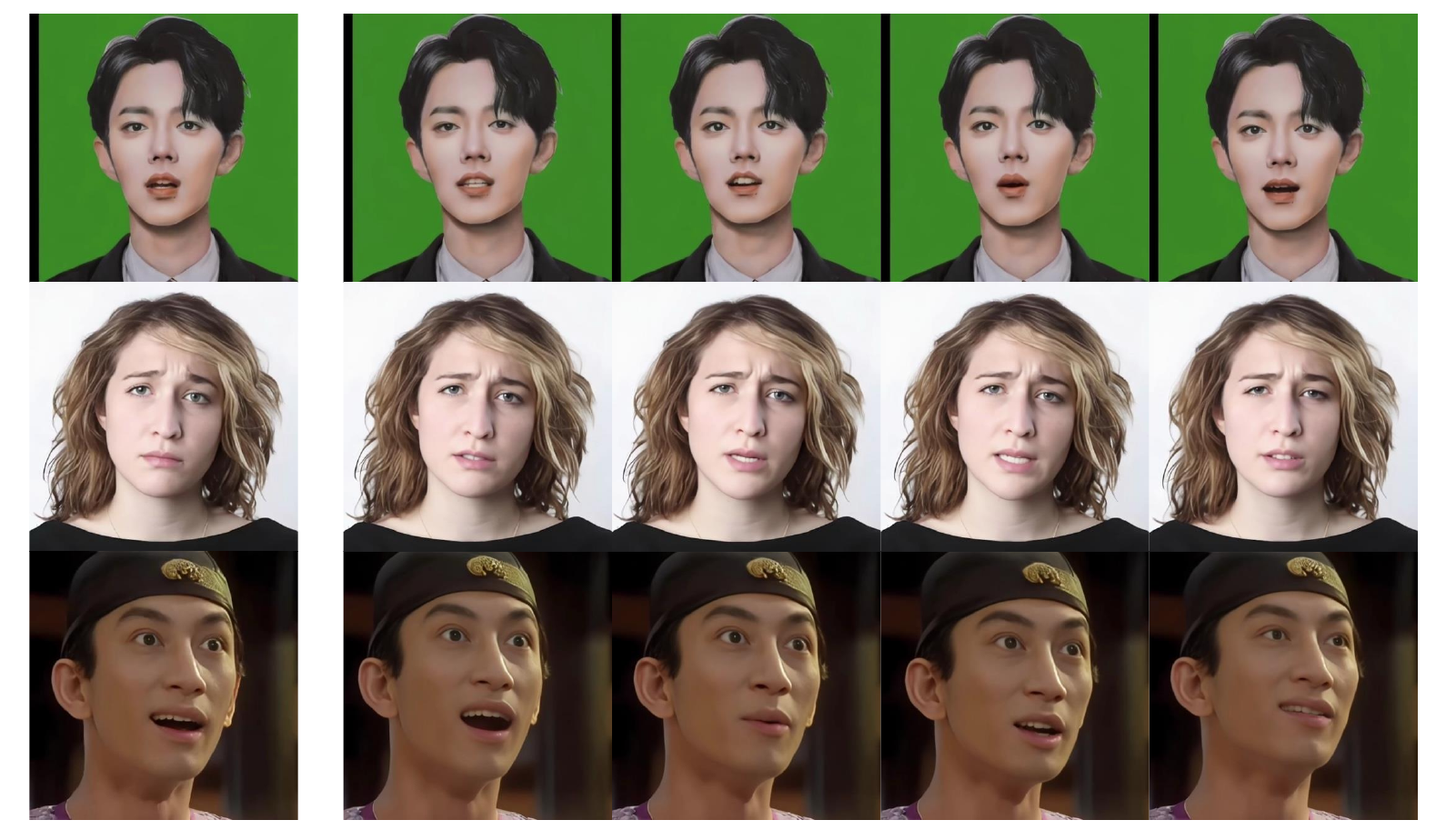}
    \caption{The qualitative results based on different avatar styles. The left image is the reference image. We sample four frames from each clip.}
    \label{demos}
\end{figure}

\begin{figure} 
    \centering
    \includegraphics[scale=0.6]{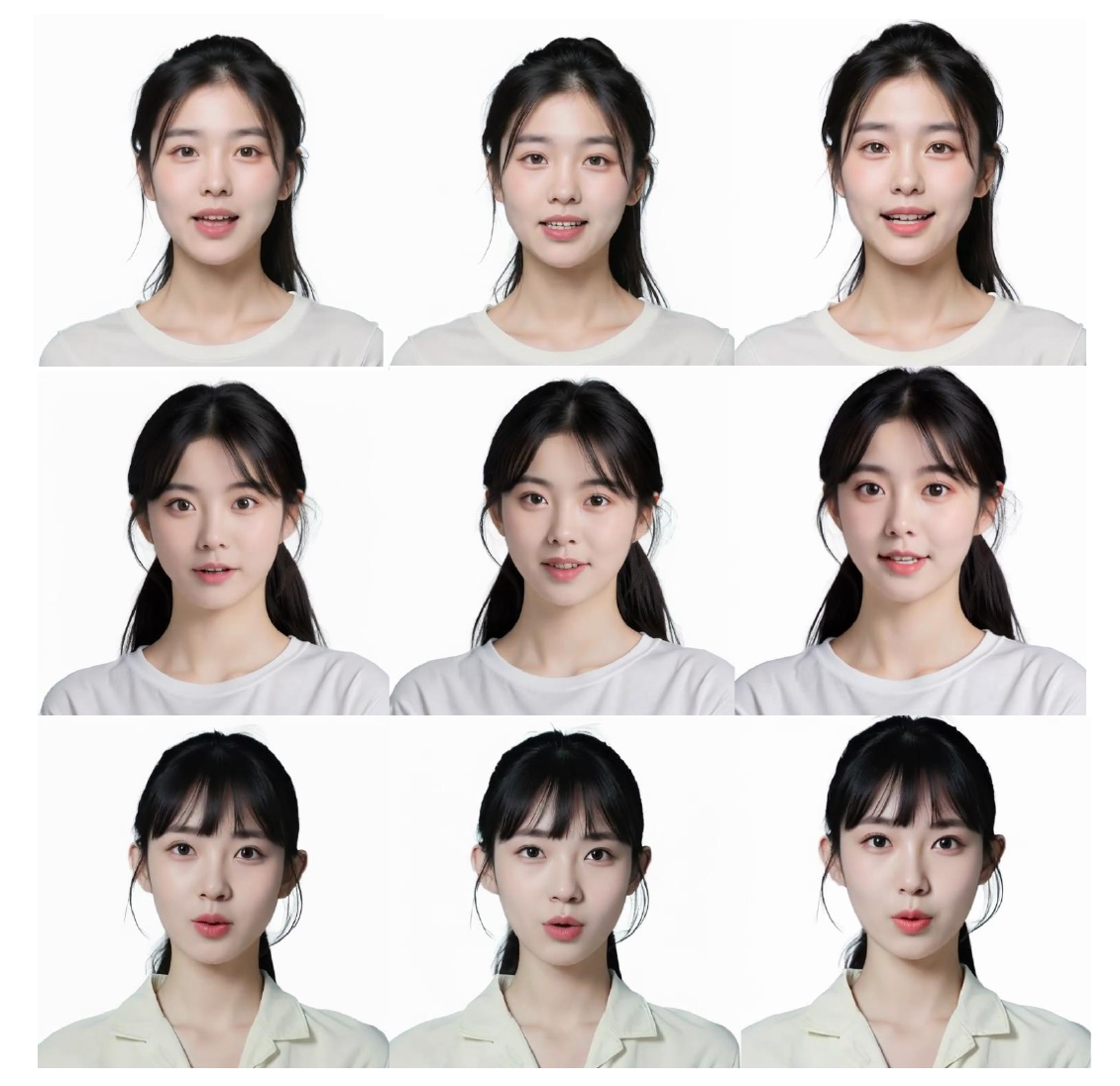}
    \caption{From left to right, the portraits are generated with the same audio segment under 2-step, 4-step, and 25-step sampling, respectively.}
    \label{lcm}
\end{figure}

\begin{figure} 
    \centering
    \includegraphics[scale=0.56]{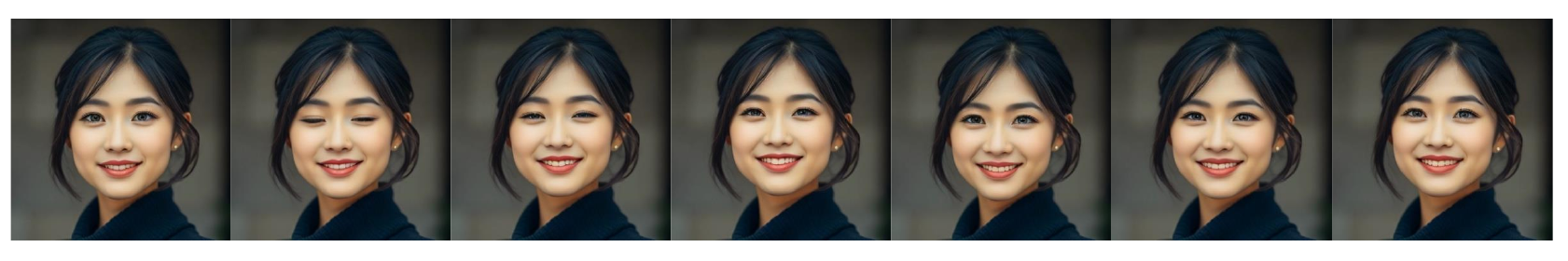}
    \caption{Illustration of the digital avatar in the listening state. The avatar performs a nodding response action according to the input class label prompt.}
    \label{listen}
\end{figure}

\begin{figure} 
    \centering
    \includegraphics[scale=0.56]{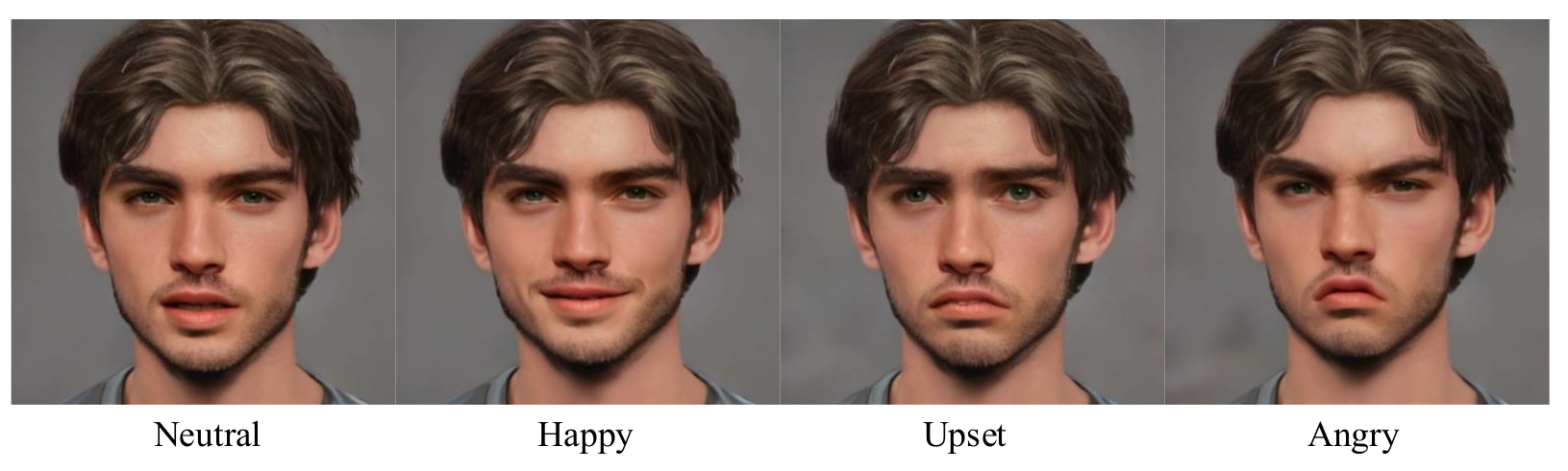}
    \caption{The qualitative results of expression control. In the generated 20-second video, the model smoothly switches between various expressions according to the input reference images with different facial expressions.}
    \label{emot}
\end{figure}

\subsection{Quantitative Results}
\paragraph{Inference Speed}
During the inference stage, each generation samples $n + f$ contiguous frames, where $n$ is fixed at 4 frames. To achieve a lower response latency, for the initial segments and during idle periods without user input, we set frames $f = 4$. In other situations, we set $f = 12$ frames to improve the quality of the generated segments. Table \ref{tab:Inference Speed} presents the inference speed of the model under various parameter configurations. In this table, the "pipe" speed refers to the total runtime of all modules except for the VAE decoder, as the decoder can run in parallel within our pipeline. "DenoisingNet" specifically indicates the runtime of the UNet alone. On an RTX 4090D GPU, the online generation of \( 384 \times 384 \) videos using 2-step sampling reaches up to 78 FPS with a response latency of 140 ms ("pipe" speed plus 40 ms for VAE decoder). For \( 512 \times 512 \) resolution, its speed reaches up to 45 FPS with a response delay of 215 ms, thereby fulfilling real-time performance requirements.

\subsection{Qualitative results}
\paragraph{Different Avatar Styles} We explore the model’s generative capabilities across different avatar styles. As shown in Figure \ref{demos}, although the model was trained only on realistic data, it can handle portraits generated by text-to-image models, and it is also robust to different viewpoints of real human subjects.

\paragraph{Few-step Generation} Figure \ref{lcm} demonstrates that, after being trained as consistency models, our model can maintain comparable generation quality to the 25-step sampling even when using only 4-step or as few as 2-step, while the 25-step generation exhibits a slight advantage in image sharpness. In addition, there is virtually no difference between the results of the 4-step and 2-step models.

\paragraph{State Switching} Figure \ref{listen} showcases our model’s capability for state switching and its performance in the listening state. By modifying the class label, our model can smoothly transition from the silent or speaking state to the listening state. In the listening state, the model exhibits responsive actions, such as nodding.

\paragraph{Expression Control} Figure \ref{emot} demonstrates the ability of expression control. We generated a 20-second video and provided four different expression templates. Every 5 seconds, we employ the portrait animation to modify the expression of the reference image. Even when there are significant differences between expressions, our model can seamlessly generate video segments corresponding to the specified expressions.

\section{Conclusion}
In this study, we proposed an audio-driven portrait video generation technique that enables low response latency, real-time interactive virtual avatars. Our research demonstrates that the digital avatars powered by diffusion models are also capable of real-time interaction with users and maintaining low response latency. Experimental results further validate the system’s outstanding performance in terms of generation quality, seamless multi-state switching, and expression control, all with extremely low latency, making it well-suited for practical interactive applications. This research opens up new possibilities and perspectives for leveraging diffusion models in time-sensitive, user-driven multimedia generation scenarios.

\bibliographystyle{unsrt}  
\bibliography{references}

\end{document}